\documentclass[10pt,twocolumn,final]{IEEEtran}
\usepackage[dvips]{graphicx}
\usepackage[para]{threeparttable}
\usepackage{amsmath}
\usepackage[noadjust]{cite}
\ifCLASSINFOpdf
\else
\fi
\begin{document}
%
\title{Registration of Images with Outliers Using Joint Saliency Map}
%
%
%

\author{Binjie~Qin,~\IEEEmembership{Member,~IEEE,}
        Zhijun~Gu,~Xianjun~Sun,~and~Yisong Lv
\thanks{This work was supported in part by NSFC (60872102), NBRPC (2010CB834300),
the Science Foundation of Shanghai Municipal Science \& Technology Commission (04JC14060), Shanghai Municipal Health Bureau (2008115)
and the small animal imaging project (06-545).}
\thanks{Binjie Qin, Zhijun Gu and Xianjun Sun are with the Department of Biomedical Engineering, School of Life Sciences \& Biotechnology, Shanghai Jiao Tong University, Shanghai, 200240, China (e-mail: bjqin@sjtu.edu.cn; gzj0126@gmail.com; sxj\_sun@sjtu.edu.cn).}
\thanks{Yisong Lv is with the Department of Mathematics, Shanghai Jiao Tong University (e-mail: yslv@sjtu.edu.cn).}
\thanks{It is preprint version for only personal use.}}

%
%

\markboth{Preprint version for IEEE SIGNAL PROCESSING LETTERS}%
{Shell \MakeLowercase{\textit{et al.}}: Bare Demo of IEEEtran.cls for Journals}
%



\maketitle

\begin{abstract}
Mutual information (MI) is a popular similarity measure for image registration, whereby good registration can be achieved by maximizing the compactness of the clusters in the joint histogram. However, MI is sensitive to the ``outlier'' objects that appear in one image but not the other, and also suffers from local and biased maxima. We propose a novel joint saliency map (JSM) to highlight the corresponding salient structures in the two images, and emphatically group those salient structures into the smoothed compact clusters in the weighted joint histogram. This strategy could solve both the outlier and the local maxima problems. Experimental results show that the JSM-MI based algorithm is not only accurate but also robust for registration of challenging image pairs with outliers.
\end{abstract}

\begin{IEEEkeywords}
image registration, mutual information, outliers, joint saliency map, weighted joint histogram.
\end{IEEEkeywords}

%

\section{Introduction}
%
%
%
%
\IEEEPARstart{I}{mage} registration can be considered as finding the optimal transformation $T$ between the reference image $I_R$ and the floating image $I_F$ to maximize a defined similarity measure such as mutual information (MI). Since 1995 \cite{1}\cite{2}, MI has been proved to be very effective in image registration. The MI between $I_R $ and $I_F $ (with intensity bins $r$ and $f$) is defined as:
\begin{equation}
\label{eq1}
\mbox{MI}=H\left({I_R }\right)+H\left({I_F}\right)-H\left({I_R,I_F}\right)
\end{equation}
where $H\left({I}\right)=-\sum_{i}p\left({i}\right)\log p\left({i}\right)$ and $H\left({I_R,I_F}\right)=-\sum_{r,f}p\left({r,f}\right)\log p\left({r,f}\right)$ are the entropy of the intensities of image $I$ and the entropy of the joint intensities of two
images, $p\left({i}\right)$ is the intensity probabilities with $p\left(r\right)=\sum\nolimits_f{p\left({r,f}\right)}$ and $p\left(f\right)=\sum\nolimits_r {p\left({r,f}\right)}$, $p\left( {r,f}\right)$ is the joint intensity probabilities estimated by the joint histogram $h\left({r,f}\right)$.

MI-based registration methods take advantage of the fact that properly registered images usually correspond to compactly-clustered joint histograms \cite{3}. They measure the joint histogram dispersion by computing the entropy of the joint intensity probabilities. When the images become misregistered, the compact clusters become disperse sets of points in the joint histogram and the entropy of the joint intensity probabilities increases. Making no assumptions about the form of the intensity mapping between the two images, MI is sensitive to the unmatchable outliers, e.g. the tumor resection in the intra- and pre-operative brain images (see Figs. 1a-b). To reject the outliers, some approaches are proposed including consistency test \cite{4}, intensity transformation \cite{5}, gradient-based asymmetric multifeature MI \cite{6} and graph-based multifeature MI \cite{7}. However, all these methods do not emphasize the corresponding salient structures in the two
images to suppress the outliers. Furthermore, MI likely suffers
from local and biased maxima \cite{8} which are caused by the ambiguities in defining
structure correspondence.
\begin{figure}[!t]
\centerline{\includegraphics[width=3.45in,height=0.80in]{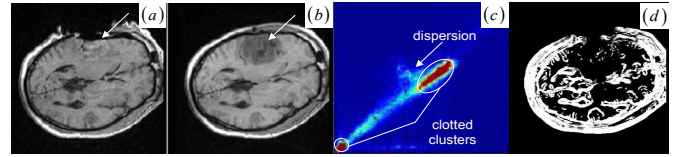}}
\caption{(a)-(b) Intra-operative and pre-operative MR image with a large tumor
resection. (c) Joint histogram dispersion with two clotted clusters (dark
red in pseudo color). (d) Joint saliency map for (a) and (b).}
\label{fig1}
\end{figure}

Spatial information, i.e. the dependence of the intensities of neighboring
pixels, has been included in MI \cite{9}-\cite{12} to improve
registration. Nevertheless, almost all MI-based methods equally
treat each overlapping pixel pair as a separate point in the overlap area to calculate the joint histogram. This could raise three
issues: 1) when we equally consider the outlier pixel pairs,
the noncorresponding structures overlap and the histogram will show certain
clusters for the grey values of the outliers. These clusters easily
introduce the histogram dispersion (see Fig. 1c) with increasing misregistration; 2) while registration can be achieved by maximizing the compactness of the histogram, the undesired clotted clusters (see Fig. 1c) related to many noisy pixel pairs in the structureless
regions, such as background and white matter in the brain image, increase the MI ambiguities and the local maxima \cite{8} (Fig. 5c shows that the normalized MI \cite{1}\cite{20} is in a biased global maximum when the whole background areas in the two endoscopic images
are exactly aligned); 3) when we group the intensity pairs as separate
points into the histogram, the independence of the neighboring bins could increase the MI ambiguities and the
local maxima. To solve this problem, joint histogram smoothing (or blurring) \cite{6}\cite{8} has been used to increase the dependence of the neighboring histogram bins. We address these issues above as follows.

In fact, image registration is to match the corresponding salient structures in both images. To suppress the outliers and the homogeneous pixel pairs, the corresponding pixel pairs in the corresponding salient structures should contribute more to the joint histogram. For example, the corresponding salient pixel pairs in the normal brain tissues should be given more weight in the histogram
than the homogeneous and the tumor resection pixel pairs. To weight each overlapping pixel pair when computing the joint histogram, we propose a novel joint saliency map (JSM) to assign a joint saliency value between 0 and 1 to the pixel pair. The idea of JSM is demonstrated schematically in Fig. 1d, where the high joint saliency values are assigned to the corresponding salient pixel pairs rather than the outlier and the homogeneous pixel pairs.

The JSM is determined by correlating each overlapping pixel pair's respective regional saliency vectors
(RSVs). The RSV characterizes the regional salient structure around each underlying pixel after a principal axis analysis (PAA) of the pixel's regional saliency distribution. In the JSM-weighted joint histogram (WJH), the contributions of the corresponding salient structures are distributed over neighboring histogram bins. This leads to the smoothing of the compact clusters for the grey values of the corresponding salient structures, which can solve both the outlier and the local maxima problems.

The proposed JSM-MI has been applied to the rigid registration of 2D images.
Experimental results show that, compared to other MI-based registration
methods, JSM-MI method achieves better robustness and higher
accuracy for the registration of challenging image pairs with outliers. The
letter is organized as follows. We first introduce the JSM for WJH in MI. Next, we report some experiment results to identify the registration performance on accuracy and robustness. Finally, the conclusions close this letter.

\section{Methods}
\subsection{Regional Saliency Vector}
We use visual saliency operator to enhance the regional salient structures we are interested in. Many techniques have been developed to define the saliency of image, i.e., using edge gradient, local phase \cite{12}, salient regions \cite{13}, corner and keypoints \cite{14}. Gradient map has been incorporated into the MI-based registration methods \cite{9}-\cite{11}. However, gradient is a local feature and sensitive to noise. Local phase \cite{12} and salient regions \cite{15} suffer from high computational complexity. Corner and keypoint can not be defined for each image pixel. Inspired by the center-surround mechanism \cite{16}\cite{17} which has defined the intensity-contrast-based visual saliency map, we define a two-step scale and rotation invariant saliency operator based on intensity contrast as follows:
\begin{equation}
\label{eq1}
S_l (v)=\sum\nolimits_{u\in N_v } {\left( {I_l \left( v \right)-I_l \left( u
\right)} \right)^2}
\end{equation}
where $N_v $ is the 1-pixel radius circular neighborhood of the pixel position $v=\left( {x,y} \right)$ at scale $l$, $S_l (v)$ is the local saliency computed for the intensity $I_l (v)$ in the Gaussian image pyramid \cite{18} at scale $l$, $I_l (u)$ is the intensity of the pixel in the $I_l (v)$'s neighborhood. The multiscale local saliency map $S(x,y)$ at the finest scale is reconstructed by summing up all the saliency maps at the coarser scales.

In the second step, a PAA of the saliency distribution in a certain region assigns \emph{regional saliency} to each pixel based on the inertia matrix:
\begin{equation}
\label{eq1}
\boldsymbol{M}=\left[ {\begin{array}{l}
 \mu _{20} \;\;\;\;\mu _{11} \\
 \mu _{11} \;\;\;\;\mu _{02} \\
 \end{array}} \right]
\end{equation}
where $\mu_{jk} =\sum\nolimits {(x-g_x )^j(y-g_y )^k} S(x,y)$ , $(g_x,g_y
)=({{m_{10} } \mathord{\left/ {\vphantom {{m_{10} } {m_{00} }}} \right.
\kern-\nulldelimiterspace} {m_{00} },m_{01} } \mathord{\left/ {\vphantom
{{{m_{10} } \mathord{\left/ {\vphantom {{m_{10} } {m_{00} }}} \right.
\kern-\nulldelimiterspace} {m_{00} },m_{01} } {m_{00} }}} \right.
\kern-\nulldelimiterspace} {m_{00} })$ and $m_{jk} =\sum\nolimits {x^jy^kS(x,y)}$ are the central $(j,k)\mbox{-}$ moment, the centroid and the $(j,k)\mbox{-}$ moment of the saliency distribution $S(x,y)$ in the 5.5-pixel radius circular neighborhood around each pixel. This regional saliency distribution describes a 2D regional salient structure. The two eigenvectors of the matrix $\boldsymbol{M}$ represent the orthogonal coordinate system within the regional salient structure, while the corresponding eigenvalues give information about the length of the respective axes. Because the regional information about the orientation of the salient structure is mostly stored along the first eigenvector corresponding to the largest eigenvalue, the first eigenvector referred as the RSV is enough to represent the regional salient structure around a pixel (see Figs. 2a-b).

\begin{figure}[!t]
\centerline{\includegraphics[width=3.45in,height=0.80in]{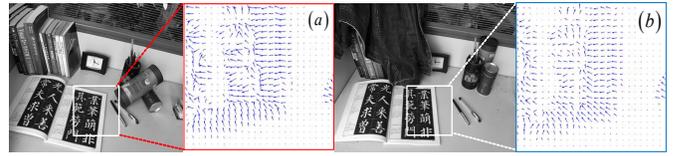}}
\caption{(a)-(b) RSVs for the sub-blocks in the reference and the floating images (size: $400\times300$ pixels).}
\label{fig2}
\end{figure}

\subsection{Joint Saliency Map}
Given two RSVs of each overlapping pixel pair, JSM is ready to describe the matching degree between the two RSVs. The inner product of two RSVs is a measure of their co-linearity and naturally can be used as their similarity
measure. The essential idea of JSM is an assumption, which
is always valid in practice according to the empirical experience in image
registration: for two precisely aligned multi-modal (or multi-temporal)
images, the majority of the corresponding pixel locations are very likely to
produce the RSVs with similar orientations (see Figs. 2a-b). This is because the
two images under registration fundamentally depict the same image
structures. As a result, the RSVs of the corresponding pixel locations from
two images could present relatively coincident orientations in general.
Therefore, the angle $\theta$ between the two RSVs (${\rm {\bf x}}_R $, ${\rm {\bf x}}_F)$ is
simply calculated, making $\cos\theta$ the scalar measure of the joint
saliency value $w\left(v \right)$:
\begin{equation}
\label{eq1}
w\left(v \right)\mbox{=}\cos \theta \left( {{\rm {\bf x}}_R ,{\rm {\bf
x}}_F } \right)={\left\langle {{\rm {\bf x}}_R ,{\rm {\bf x}}_F }
\right\rangle } \mathord{\left/ {\vphantom {{\left\langle {{\rm {\bf x}}_R
,{\rm {\bf x}}_F } \right\rangle } {\left\| {{\rm {\bf x}}_R } \right\|\cdot
\left\| {{\rm {\bf x}}_F } \right\|}}} \right. \kern-\nulldelimiterspace}
{\left\| {{\rm {\bf x}}_R } \right\|\cdot \left\| {{\rm {\bf x}}_F }
\right\|}
\end{equation}

A JSM value near one suggests that the underlying pixel pair originates from the corresponding salient
structures. Contrarily, a JSM value near zero indicates that the underlying pixel pair comes from either the outliers or a homogeneous region. To speed up the registration without reducing accuracy, the pixel with a small saliency value below a threshold value (10 percent of the maximum saliency value) is assigned a zero JSM value directly. The JSM would primarily respond to the high-gradient edge pixels if a high threshold value is chosen. However, the JSM does not simply emphasize the common image gradients in the two images. Figs. 3d-f present the image gradient and the JSM profiles of the same line (marked as dashed lines across the tumor areas) at the two registered images (see Figs. 3a-b). As shown in the figures, the image gradient features in Figs. 3d-e are very noisy and do not agree with each other at each overlapping location. The JSM in Fig. 3f can accurately preserve the corresponding salient structures in larger capture range with smaller variability than the image gradients.

\begin{figure}[!t]
\centerline{\includegraphics[width=2.44in,height=1.60in]{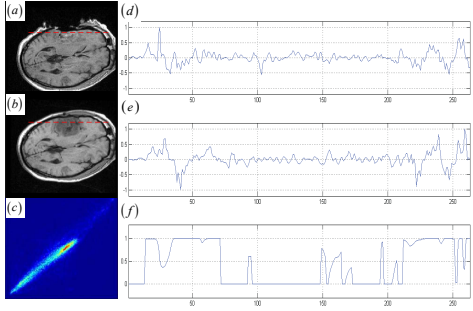}}
\caption{(a)-(b) The reference and the floating images for the gradient magnitude
and the JSM magnitude. (c) Compact JSM-WJH smoothing for (a)-(b). (d)-(e)
Gradient value profiles of the lines in (a)-(b), which are marked as dashed
lines. (f) JSM value profiles of the lines in (a)-(b).}
\label{fig3}
\end{figure}

\subsection{JSM-Weighted Joint Histogram }
The contribution of the interpolated floating intensity $f(v_f)$ to the joint histogram is weighted by a $w(v)$ of the JSM (the pixel positions ($v_r$,$v_f$) are overlapped at the position $v$). For 2D image registration, if using a nearest neighbor or a bilinear interpolation, the value $w(v)$ should be added to the histogram entry $h(r,f)$. In bilinear partial volume distribution (PV) interpolation, the contribution of the $f(v_f)$ to the histogram, distributed over the intensity values of all nearest neighbors of the reference pixel position $v_r$ on the grid of $I_R$, is weighted using the $w(v)$. Similarly, JSM could be easily incorporated into other interpolation schemes and Parzen-based joint histogram.

In the JSM-WJH, the outliers and homogeneous regions have little impact on the histogram distribution.
Furthermore, each histogram entry for the corresponding salient structures is the sum of smoothly varying
fractions of one, such that the histogram changes smoothly in the neighboring bins related to those structures. As a result, the compact histogram smoothing (see Fig. 3c) is introduced by highlighting the grey values of the corresponding salient structures. Computed from the compact and smooth histogram, the MI is then maximized to achieve robust and accurate rigid registration.

\subsection{Computational complexity}
The JSM should be re-calculated with the transformation changing the overlap area at each registration iteration. The RSV orientation for a JSM calculation could be easily re-oriented as it is done in the diffusion tensor image registration \cite{19}. Nevertheless, to ensure the numerical stability and the computation speedup, a new JSM at each iteration can be simply updated from the JSM of the previous iteration through the PV interpolation. The JSM could be re-calculated after $n$ iterations ($n{=}10\sim15$) to reflect the updated correspondence between the salient structures in the two images.

\section{Experimental results}
We evaluated our JSM-MI-based (JMI) algorithm on 11 challenging image pairs including CT-PET tumor images, MR brain tumor resection images, optical images with background/foreground clutter and etc. We implemented the JMI algorithm using the simplex optimization in a multiresolution scheme \cite{18}. The algorithm stops if the current step length is smaller than ${10^{-5}}$ or if it has reached the limit of 200 evaluation numbers. The challenging image pairs include some complex outliers that the normalized MI-based method and four of MI-based adaptations with incorporating spatial information fail to deal with. Due to space restrictions, we only show some typical experimental results in this letter.

\begin{figure}[!t]
\centerline{\includegraphics[width=3.44in,height=2.5in]{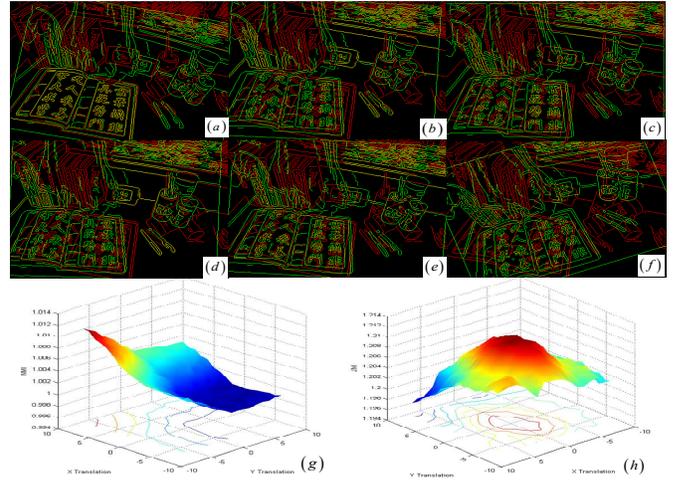}}
\caption{Registration results for the two images in Figs. 2a-b. (a) JMI. The yellow contour overlap of the book validates the registration accuracy owing
to the additive mixing of red and green. (b) NMI. (c) RMI. (d)
HMI. (e) GMI. (f) PMI. (g)-(h) NMI and JMI similarity surfaces plotted as a function of $x$ and $y$ translation (within a range of $\pm10$ pixels around the matching position)}
\label{fig4}
\end{figure}

Fig. 4 shows the various registration results for
the two images at Fig. 2 with a foreground book and the large changes of background appearance. To facilitate the visual assessment of registration accuracy,
the green floating contours and the red reference contours obtained by Canny-Deriche edge detector have been overlaid
over each other. The sub-pixel registration accuracy (see Table I. case 1)
of our JMI algorithm can be validated by the book's yellow contour overlap, which is due to the additive color mixing of
the green and the red contour (see Fig. 4a).

Using particle swarm optimization (PSO) to deal with the local maxima,
the other methods based on normalized MI (NMI)
\cite{1}\cite{20}, regional MI (RMI)
\cite{21}, high-dimensional MI (HMI)
\cite{22}, MI with gradient information (GMI)
\cite{10}, and phase MI (PMI)
\cite{12} show different misregistration results in Figs. 4b-f. The PSO is conducted with 20 particles and allowed to experience 2000 iterations. The algorithm stops if it has reached the limit of 200 evaluation numbers or if the minimum error (${10^{-5}}$) conditions is satisfied. The computation time needed for the different algorithms are listed in Table II.

Figs. 4g-h show that the NMI and JMI similarity surfaces are plotted as a function of $x$ and $y$ translation. In this case, the JSM removes all local maxima and achieves the global maximum at the registration position, while the NMI suffers from the biased maximum at the mismatching position.

Figs. 5a-b show the reference and floating endoscopic images ($720\times572$ pixels) including a surgical instrument with different illuminations. Using a mosaic pattern to fuse the two images, Figs. 5c-d show the NMI-based and PMI-based misregistration results. Fig. 5e shows our accurate JMI-based registration result (see Table I. case 2).

\begin{figure}[!t]
\centerline{\includegraphics[width=2.2in,height=1.5in]{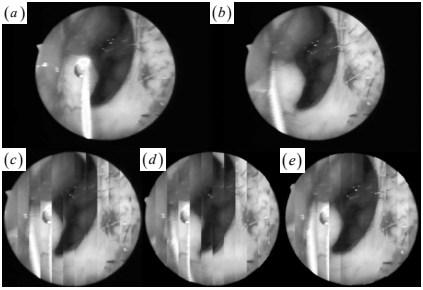}}
\caption{(a)-(b) Reference and floating endoscopic images (size: $720\times572$ pixels) with a
surgical tool and illumination changes. The two images are fused using a mosaic pattern. (c) NMI. (d)
PMI. (e) JMI.}
\label{fig5}
\end{figure}

\begin{center}
\begin{threeparttable}
\caption{Registration results for Fig. 4 and Fig. 5 (The translations $X$ and $Y$ are in pixels in the $x$ and $y$ directions,
the rotation $\beta$ is in degrees around the center of the images.).}
\begin{tabular*}{0.48\textwidth}{@{\extracolsep{\fill}} rlcc }
\hline
&Cases&Correct($X$,$Y$,$\beta$)&Computed($X$,$Y$,$\beta$)\\
\hline
&1&$-23.11$, $45.59$, $11.43^\circ$&$-22.34$, $45.30$, $11.03^\circ$\\
&2&$37.91$, $-36.78$, $4.43^\circ$&$37.46$, $-38.18$, $4.68^\circ$\\
\hline
\end{tabular*}
\end{threeparttable}
\end{center}

\begin{center}
\begin{threeparttable}
\caption{Computation iterations and runtime in seconds for Fig. 4. (Matlab 6.5, single core Intel Celeron 2.8GHz, RAM 2GB)}
\begin{tabular*}{0.48\textwidth}{@{\extracolsep{\fill}}rlcccccc}
\hline
&&JMI&NMI&RMI&HMI&GMI&PMI\\
\hline
&Iter.&64&41&45&46&50&29\\
&Time&157.4&296.7&297.1&1060.1&329.1&3049.3\\
\hline
\end{tabular*}
\end{threeparttable}
\end{center}

\section{Conclusion}
We propose an effective JSM to solve the problems of outliers and local maxima in MI-based
image registration. Representing the corresponding salient structures in the two images to be registered, JSM is easily integrated into other intensity-based similarity measures for 3D nonrigid registration. Independent of this work but subsequent to our preliminary
conference papers \cite{23}\cite{24} which this letter elaborates on and extends, Ou et al. \cite{25} developed a similar mutual saliency map for
outlier rejection in 3D nonrigid image registration.

Additionally, our method is an intensity-based method and also sensitive to the initial conditions. It is necessary in principle to set the proper initial conditions close to a correct alignment solution, which can be achieved by coarse alignment techniques such as principal axes based method. Nevertheless, all instances of correct registration in this letter are directly performed by our method without any coarse alignment.


%



\section*{Acknowledgment}

The authors thank Simon K. Warfield, Michal Irani, Robert Barnett and Edward Vrscay for allowing the use of image, Rehan Ali for the phase recovery source code, Shanbao Tong and all reviewers for their useful comments, and Wendy Wang for her help to our algorithm.

\ifCLASSOPTIONcaptionsoff
  \newpage
\fi

\end{document}